\documentclass[10pt]{article}
\usepackage{spconf,amsmath,graphicx}

\usepackage[font=footnotesize]{caption}

\newif\ifarxiv
\newif\ifnotarxiv

\arxivtrue
\notarxivfalse

\usepackage{palatino}

\usepackage{amsmath,graphicx}

\usepackage[english]{babel}
\usepackage[T1]{fontenc}
\usepackage{amsfonts, amssymb}
\usepackage{setspace}
\usepackage{pifont}
\usepackage[normalem]{ulem}
\usepackage{placeins}
\usepackage{xspace}

\usepackage{dblfloatfix}    %

\usepackage[table,dvipsnames]{xcolor}

\newcommand{\matthijs}[1]{{\color{RedViolet} [\textbf{Matthijs}: #1]}}

\newcommand{\pf}[1]{{\color{Aquamarine} [\textbf{Pierre}: #1]}}

\usepackage[colorlinks=true, allcolors=blue]{hyperref}
\usepackage{subcaption} %
\usepackage{algorithm, algpseudocode}

\usepackage{booktabs}

\newcommand{\R}{\mathbb{R}}

\newcommand{\abs}[1]{\lvert{#1}\rvert}
\newcommand{\norm}[1]{\lVert{#1}\rVert}
\newcommand{\T}{^\intercal}

\newcommand{\head}[1]{\noindent \vspace{-0.75em}\\ \textbf{#1.}}
\newcommand{\headbis}[1]{\textbf{#1}}
\newcommand{\etal}{{et al}.\@ }
\newcommand{\eg}{{e.g}.\@ }

\newcommand{\aka}{{a.k.a}.\@ }

\newcommand{\soutx}[1]{}

\title{Watermarking Images in Self-Supervised Latent Spaces}

\name{
Pierre Fernandez$^{1,2}$,
Alexandre Sablayrolles$^1$,
Teddy Furon$^2$,
Herv\'e J\'egou$^1$,
Matthijs Douze$^1$
}

\address{$^1$Meta AI, $^2$Inria
\sthanks{Univ. Rennes, Inria, CNRS, IRISA. \newline Correspondence to: pfz@fb.com}}

\begin{document}

\ninept

\maketitle

\begin{abstract}

We revisit watermarking techniques based on pre-trained deep networks, in the light of self-supervised approaches.
We present a way to embed both marks and binary messages into their latent spaces, leveraging data augmentation at marking time. 
Our method can operate at any resolution and creates watermarks robust to a broad range of transformations (rotations, crops, JPEG, contrast, etc). 
It significantly outperforms the previous zero-bit methods, and its performance on multi-bit watermarking is on par with state-of-the-art encoder-decoder architectures trained end-to-end for watermarking.
The code is available at \href{https://github.com/facebookresearch/ssl\_watermarking}{github.com/facebookresearch/ssl\_watermarking}
\end{abstract}

\section{Introduction}\label{sec:introduction}

\ifarxiv
\begin{figure*}[t]
    \centering
    \vspace{-0.7cm}
    \includegraphics[width=0.9\textwidth]{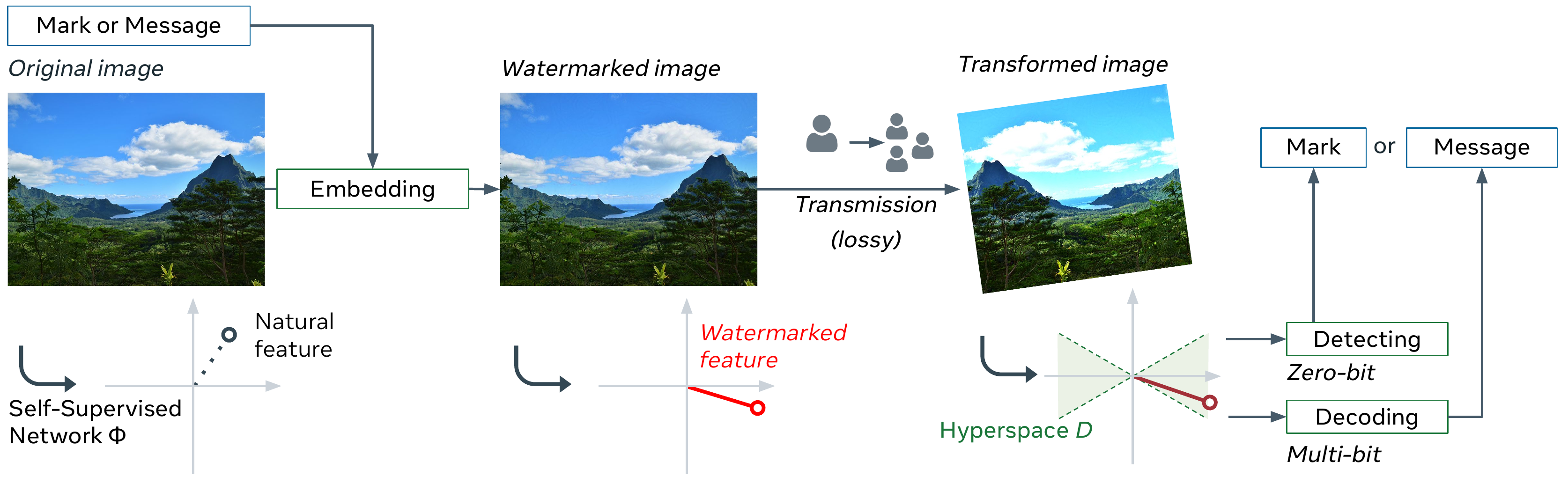}
    \caption{Overview of our method for watermarking in self-supervised-latent spaces. 
    A self-supervised network trained with DINO \cite{caron2021dino} builds a latent space on which the mark is added by the embedding process. 
    Its effect is to shift the image's feature into a well-specified region of the latent space, such that transformations applied during transmission do not move much the feature. The mark's detection (zero-bit watermarking setup) or message's decoding (multi-bit watermarking setup) is performed in the same latent space.}
    \label{fig:splash}
\end{figure*}
\fi

Watermarking embeds a secret message into an image under the constraints of 
(1) \emph{imperceptibility} -- the distortion induced by the watermark must be invisible, %
(2) \emph{payload} -- the hidden message is a binary word of a given length, 
(3) \emph{robustness} -- the decoder retrieves the hidden message even if the image has been distorted to some extent.
This paper deals with blind watermarking where the decoder does not have access to the original image.
The classic approach, coined \emph{TEmIt} (Transform, Embed, Inverse transform) by T. Kalker, embeds the watermark signal in the feature space of a transform (\eg DFT, DCT, Wavelet). It provides \emph{perceptually significant coefficients} reliable for watermarking as conceptualized in~\cite[Sec.~8.1.3]{cox2007digital}. 

Watermarking is enjoying renewed interest from advancements in deep learning. 
New methods improve the robustness to a broad range of alterations thanks to neural networks offering a reliable latent space where to embed the information.
Examples include directly marking into the semantic space resulting from a supervised training over a given set of classes like ImageNet~\cite{vukotic2020classification}, or explicitly training a watermarking network to be invariant to a set of image perturbations. 
In this case, networks are usually encoder-decoder architectures trained end-to-end for watermarking~\cite{zhu2018hidden,wen2019romark, ahmadi2020redmark, zhang2020robust}.

Our key insight is to leverage the properties of \emph{self-supervised} networks to watermark images.
Ideally, according to~\cite{cox2007digital}, a perceptually significant coefficient does not change unless the visual content of the image is different.
Similarly, some self-supervised methods aim to create representations invariant to augmentations, without explicit knowledge of the image semantics~\cite{caron2021dino,grill2020byol}.
These pre-trained networks offer us the desired embedding space "for free", saving us the heavy training of end-to-end architectures like HiDDeN~\cite{zhu2018hidden}.

In order to robustly embed in the latent spaces, gradient descent is performed over the pixels of the images. 
To further ensure both robustness and imperceptibility of the watermarks, we include data augmentation and image pre-processing at marking time.

\ifnotarxiv
\noindent
\fi
Our contributions are the following:
\begin{itemize}
    \setlength\itemsep{0.1em}
    \item We provide a watermarking algorithm that can encode both marks and binary messages in the latent spaces of any pre-trained network;
    \item We leverage data augmentation at marking time;
    \item We experimentally show that networks trained with self-supervision provide excellent embedding spaces.
\end{itemize}

\section{Related work} \label{sec:relatedwork}

\noindent
\headbis{Image watermarking}~\cite{cox2007digital} 
approaches are often classified by their embedding space: few use the spatial domain
\cite{ni2006reversible,nikolaidis1998robust}, most of them follow the TEmIt principle 
with a well known transformation like DFT~\cite{urvoy2014perceptual}, DCT~\cite{bors1996image} or DWT~\cite{xia1998wavelet}.
In \textbf{zero-bit watermarking}~\cite{furon2007constructive}, the embedder only hides a mark and the \emph{detector} checks for its presence in the content. 
For instance, the received content is deemed watermarked if its descriptor lies in a hypercone of the feature space. 
This detection strategy is near optimal~\cite{furon2019dualhypercone,merhav2008optimal}. 
In \textbf{multi-bit watermarking}, the embedder encodes a binary word into a signal that is hidden in the image. At the other end, the \emph{decoder} retrieves the hidden message bit by bit.
This is the case for most deep-learning-based methods presented below.

\vspace{0.5em}
\noindent
\headbis{Deep-learning-based watermarking}
\label{par:deep_watermarking}
has emerged as a viable alternative to traditional methods. %
The networks are often built as encoder-decoder architectures~\cite{zhu2018hidden, kandi2017cnn}, where an encoder embeds a message in the image and a decoder tries to extract it. For instance, HiDDeN \cite{zhu2018hidden} jointly trains encoder and decoder networks with noise layers that simulate image perturbations. %
It uses an adversarial discriminator to improve visual quality, and was extended to arbitrary image resolutions and message lengths by Lee \etal\cite{lee2020resolution}.
Distortion Agnostic~\cite{luo2020distortion} adds adversarial training that brings robustness to unknown transformations and~\cite{zhang2020robust, yu2020attention}  embed the mark with an attention filter further improving imperceptibility.
ReDMark~\cite{ahmadi2020redmark} adds a circular convolutional layer that diffuses the watermark signal all over the image. 
Finally, ROMark \cite{wen2019romark} uses robust optimization with worst-case attack as if an adversary were trying to remove the mark.
For more details, we refer to the review~\cite{byrnes2021survey}. 
This type of methods has its drawbacks: \eg it is hard to control the embedding distortion and it is made for decoding and does not translate so well to detection.

Vukoti\'{c} \etal\cite{vukotic2018deep} mark images with a neural network pre-trained on supervised classification instead of an encoder-decoder architecture. 
The network plays the role of the transform in a TEmIt approach. 
Since it has no explicit inverse, a gradient descent to the image pixels "pushes" the image feature vector into a hypercone. 
The follow-up work~\cite{vukotic2020classification} increases the inherent robustness of the network by applying increasingly harder data augmentation at pre-training. 
It offers a guarantee on the false positive rate without requiring to train a network explicitly for watermarking, but no multi-bit version was proposed.

\section{Watermarking with SSL networks}\label{section:0bit}

Our method adopts the framework of Vukotić \etal\cite{vukotic2020classification}. 
We improve it by showing that networks build better watermarking features when trained with self-supervision, and by introducing data-augmentation at marking time. 
We also extend it to multi-bit watermarking.

\vspace{-0.5em}
\subsection{Using self-supervised networks as feature extractors}\label{subsec:ssl_as_feature_extractor}
\vspace{-0.5em}

\def \real{\mathbb{R}}
\def \Io{I_o}
\def \Iw{I_w}
\def \Ispace{\mathcal{I}}
\def \Fspace{\mathcal{F}}
\def \Lw{\mathcal{L}_w}
\def \Li{\mathcal{L}_i}
\def \L{\mathcal{L}}
\def \Tr{\mathrm{Tr}}

\head{Motivation}
We denote the image space by $\Ispace$ and the feature space by $\Fspace=\real^d$.
The feature extractor $\phi:\Ispace\to\Fspace$ must satisfy two properties: 
(1) geometric and valuemetric transformations on image $I$ should have minimal impact on feature $\phi(I)$,
(2) it should be possible to alter $\phi(I)$ by applying invisible perturbations to the image in order to embed the watermark signal.

We choose $\phi$ as a neural network pre-trained by self-supervised learning (SSL).
Our assumption is that SSL produces excellent marking spaces because 
(1) it explicitly trains features to be invariant to data augmentation; and 
(2) it does not suffer from the \textit{semantic collapse} of supervised classification, that gets rid off any information that is not necessary to assign classes~\cite{doersch2020crosstransformers}. 
From the recent SSL methods of the litterature (contrastive learning~\cite{chen2020simclr,he2020moco}, statistical prior~\cite{zbontar2021barlow}, teacher/ student architecture~\cite{grill2020byol}), we select DINO~\cite{caron2021dino} for its training speed and content-based image retrieval performance.

\noindent
\head{DINO pre-training}
DINO~\cite{caron2021dino} pertains to the teacher/ student approach.
The teacher and the student share the same architecture.
Self distillation with no labels trains the student network to match the outputs of the teacher
on different views of the same image. 
The student is updated by gradient descent while the teacher's parameters are updated as an exponential moving average of the student's parameters: $\theta_{t} \leftarrow \lambda \theta_{t} + (1-\lambda) \theta_{s}$, with $\lambda\lesssim 1$.

The invariance of the features is ensured by random augmentations during the training: valuemetric (color jittering, Gaussian blur, solarization) and geometric (resized crops) transformations.
Furthermore, DINO encourages local-to-global correspondence by feeding only global views to the teacher while the student sees smaller crops. 

\iffalse
There is a link between the contrastive loss and the entropy of the latent space~\cite{boudiaf2020unifying, elnouby2021retrieval}: 
the positive term minimizes the conditional differential entropy between samples of the same category, 
the negative term maximizes the entropy of the learned representations.
Wang \etal\cite{wang2020uniformity} show that contrastive learning optimizes both the uniformity of features on the hypersphere.
This has great importance in watermarking, as explained in~\autoref{subsec:ssl_as_feature_extractor}.
\fi

\noindent
\head{Normalization layer}\label{par:whitening} 
The watermark embedding must drive the feature to an arbitrary space region (defined by a secret key and the message to hide).
It is essential that the features are not concentrated onto a manifold far away from this region. 
Although Wang \etal\cite{wang2020uniformity} show that contrastive learning optimizes the uniformity of the features on the hypersphere, it does not occur in practice with DINO.
To alleviate the issue, the output features are transformed by PCA-whitening (\aka PCA-sphering). 
This learned linear transformation~\cite{jegou2012negative} outputs centered vectors with unit covariance of dimension $d=2048$. 

\vspace{-0.5em}
\subsection{Marking with back-propagation and augmentation}\label{subsec:embedding_algorithm}
\vspace{-0.5em}

The marking takes an original image $\Io$ and outputs a visually similar image $\Iw$.
In the image space $\Ispace$, the distortion is measured by $\Li:\Ispace\times\Ispace\to\real_+$. An example is the MSE:
$\Li(\Iw,\Io) = \|\Iw-\Io\|^2 / h/w$, but it could be replaced by perceptual image losses such as LPIPS~\cite{zhang2018unreasonable}.

In the feature space $\Fspace$, we define a region $\mathcal{D}$ which depends on a secret key (zero-bit and multi-bit setups) and the message to be hidden (only in multi-bit setup). 
Its definition is deferred to Sect.~\ref{sec:modulation} together with the loss $\Lw:\Fspace\to\real$ that captures how far away a feature $x\in\Fspace$ lies from $\mathcal{D}$. %
We also define a set $\mathcal T$ of augmentations, which include rotation, crops, blur, etc., each with a range of parameters.
$\Tr(I,t)\in\Ispace$ denotes the application of transformation $t\in\mathcal T$ to image $I$.

If the feature extractor is perfectly invariant, $\phi(\Tr(I,t))$ lies inside $\mathcal{D}$ if $\phi(I)$ does. 
To ensure this, the watermarking uses data augmentation.
The losses $\Lw$ and $\Li$ are combined in:
\begin{equation}
\label{eq:global_loss}
    \L(I, \Io, t) := \lambda \Lw(\phi(\Tr(I,t))) + \Li(I,\Io).
\end{equation}

The term $\Lw$ aims to push the feature of \emph{any transformation} of $\Iw$ into $\mathcal D$, while the
term $\Li$ favors low distortion.
The training approach is typical for the adversarial attacks literature \cite{goodfellow2014adversarial, szegedy2013intriguing}: 
\begin{equation}
\label{eq:Watermarking}
    \Iw := \arg \min_{I\in \mathcal{C}(\Io)}
    \mathbb{E}_{t\sim \mathcal T} [\L(I, \Io, t)]
\end{equation}
where $\mathcal{C}(\Io)\subset\Ispace$ is the set of admissible images w.r.t. the original one. 
It is defined by two steps of normalization applied to the pixel-wise difference $\delta = I - \Io$:
(1) we apply a SSIM~\cite{wang2004ssim} heatmap attenuation, which scales $\delta$ pixel-wise to hide the information in perceptually less visible areas of the image;
(2) we set a minimum target PSNR and rescale $\delta$ if this target is exceeded.

The minimization is performed by stochastic gradient descent since the quality constraints, $\Tr$ and $\phi$ are differentiable w.r.t. the pixel values.
Stochasticity comes from the fact that
expectation $\mathbb{E}_{t\sim \mathcal T}$ is approximated by sampling according to a distribution over $\mathcal T$.
The final image is the rounded version of the update after $K$ iterations.

\vspace{-0.5em}
\subsection{Detection and decoding}\label{sec:modulation} 
\vspace{-0.5em}

We consider two scenarios: zero-bit (detection only) and multi-bit watermarking (decoding the hidden message).  
Contrary to HiDDeN~\cite{zhu2018hidden} and followers, our decoding is mathematically sound.

\head{Zero-bit}\label{hd:zero}
From a secret key $a\in\Fspace$ s.t. $\|a\|=1$, the detection region is the dual hypercone:
\vspace{-0.2em}
\begin{equation}
     \mathcal{D} := \left\{  x \in \R^d :  \abs{ x\T  a} > \norm{ x} \cos (\theta) \right\}.
\vspace{-0.2em}
\end{equation}
It is well grounded because the False Positive Rate (FPR) is given by
\vspace{-0.2em}
\begin{align}
   \mathrm{FPR} :=&\; \mathbb{P}\left(\phi(I)\in \mathcal D \mid \text{"key } a \text{ uniformly distributed"}\right)\nonumber\\
            =&\; 1-I_{\cos^2(\theta)} \left(\frac{1}{2}, \frac{d-1}{2} \right)
   \label{eq:FPR}
\end{align}
\vspace{-0.8em}\\
where $I_\tau (\alpha, \beta)$ is the regularized Beta incomplete function. %
Moreover, the best embedding is obtained by increasing the following function under the distortion constraint:
\vspace{-0.2em}
\begin{equation}
\vspace{-0.2em}
   - \Lw(x) = (x^\top a)^2 -  \|x\|^2 \cos^2\theta.
\end{equation}
This quantity is negative when $x\notin\mathcal{D}$ and positive otherwise.
I.~Cox \etal originally called it the robustness estimate~\cite[Sec. 5.1.3]{cox2007digital}. 
This hypercone detector is optimal under the asymptotical Gaussian setup in~\cite{furon2019dualhypercone, merhav2008optimal}.

\head{Multi-bit}
We now assume that the message to be hidden is  $m = (m_1, ..., m_k) \in \{-1,1\}^k$. 
The decoder retrieves $\hat{m}=D(I)$.
Here, the secret key is a randomly sampled orthogonal family of carriers $a_1,...., a_k \in \R^d$. 
We modulate $m$ into the signs of the projection of the feature $\phi(I)$ against each of the carriers, so the decoder is: 
\ifnotarxiv
$D(I) = [\mathrm{sign}\left(\phi(I)^\top a_1\right), ..., \mathrm{sign}\left(\phi(I)^\top a_k\right)]$.
\fi
\ifarxiv
\begin{equation}
D(I) = \left[\mathrm{sign}\left(\phi(I)^\top a_1\right), ..., \mathrm{sign}\left(\phi(I)^\top a_k\right)\right].
\end{equation}
\fi

At marking time,  the functional is now defined as the hinge loss with margin $\mu\geq 0$ on the projections: 
\vspace{-0.5em}
\begin{equation}\label{eq:loss_multibit}
    \Lw(x) = \frac{1}{k} \sum_{i=1}^k \max \left( 0, \mu - (x^\top a_i).m_i \right).
\end{equation}

\ifarxiv
\vspace{-0.5em}
\subsection{Overview of the watermarking algorithm}
\vspace{-0.5em}

\begin{algorithm}
\caption{Watermarking algorithm}\label{alg:0bit}
\begin{algorithmic}
\item \textbf{Inputs}: $I_0 \in \mathcal{I}$, targeted PSNR in dB
    \item \quad if zero-bit: $ \textrm{FPR} \in [0,1]$, $a \in \R^d$
    \item \quad if multi-bit: $m \in \{-1,1\}^k$, $(a_i)_{i=1...k} \in \R^{k\times d}$
\item \textbf{Embedding}: 
    \item \quad if zero-bit: compute $\theta (\textrm{FPR})$
    \item \quad $I_w \gets I_0$
    \item \quad For $i=1, ..., n_{iter}$:
    \item \quad \quad $I_w \xleftarrow{\textrm{constraints}} I_w$ \Comment{impose constraints}
    \item \quad \quad $I_w \xleftarrow{t \sim \mathcal T} \textrm{Tr}(I_w, t)$ \Comment{sample \& apply augmentation}
    \item \quad \quad $x \gets \phi (I_w)$ \Comment{get feature of the image}
    \item \quad \quad $\mathcal{L} \gets \lambda \mathcal L_w(x) + \mathcal L_i(I_w, I_0) $ \Comment{compute loss}
    \item \quad \quad $I_w \gets I_w + \eta \times \mathrm{Adam}(\mathcal{L})$ \Comment{update the image}
    \item \quad Return $I_w$
\item \textbf{Detecting}: 
    \item \quad $x \gets \phi(I_m)$
    \item \quad if zero-bit:
    \item \quad \quad Detect if $x\in \mathcal D \Longleftrightarrow (x^Ta)^2-\norm{x}^2\cos^2\theta >0$
    \item \quad if multi-bit:
    \item \quad \quad Return $\left[\mathrm{sign}\left(x^\top a_1\right), ..., \mathrm{sign}\left(x^\top a_k\right)\right]$
\end{algorithmic}
\end{algorithm}
\vspace{0.5cm}

\fi

\newcommand{\cmark}{\textcolor{blue!50}{\ding{51}}\xspace}%
\newcommand{\cmarkg}{\textcolor{red!50}{\ding{51}}\xspace}%
\newcommand{\xmarkg}{\ding{55}\xspace}%
\newcommand{\xmark}{\textcolor{red!50}{\ding{55}}\xspace}%
\newcommand{\rot}[1]{\rotatebox{80}{#1}}%

\section{Experiments \& results}
\label{sec:experiments}

\subsection{Experimental setup and implementation details}\label{sec:exp_setup}
\vspace{-0.5em}

\head{Data}
We evaluate our method on: 
1000 images of YFCC100M dataset~\cite{thomee2016yfcc100m} which is selected for the variety of its content, CLIC~\cite{2018clic} composed of 118 professional high-resolution images when comparing to~\cite{vukotic2020classification}, and 1000 images of MS-COCO~\cite{lin2014coco} composed of smaller images for comparison with~\cite{zhu2018hidden,luo2020distortion}.

\head{Backbone pre-training}\label{par:backbone_pretraining} 
We use the ResNet-50 \cite{he2016deep} architecture as backbone model to extract features from its last convolutional layer ($d=2048$).
It is trained on ILSVRC2012~\cite{deng2009imagenet} without labels, using 200 epochs of DINO self-supervised learning with the default parameters~\cite{caron2021dino} and with rotation augmentation.
Models trained for classification come from the torchvision library~\cite{marcel2010torchvision}. 
The PCA-whitening is learned on 100k distinct images from YFCC (resp. COCO) when evaluating on YFCC and CLIC (resp. COCO).

\ifarxiv
\begin{figure*}[b]
    \centering
    \begin{subfigure}[b]{0.95\linewidth}
        \includegraphics[width=\linewidth, trim={0 2.3em 0 0em},clip]{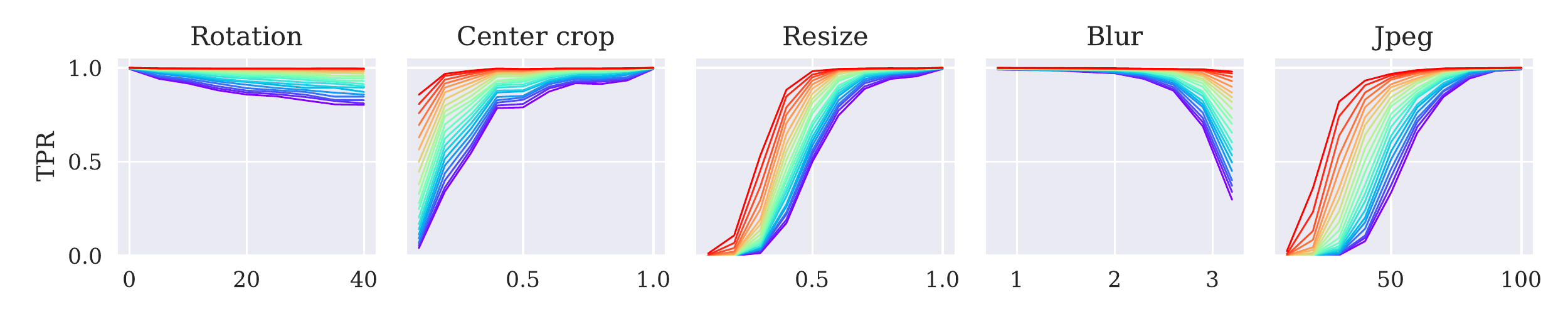}
    \end{subfigure}
    \begin{subfigure}[b]{0.95\linewidth}
        \includegraphics[width=\linewidth, trim={0 0em 0 2.7em},clip]{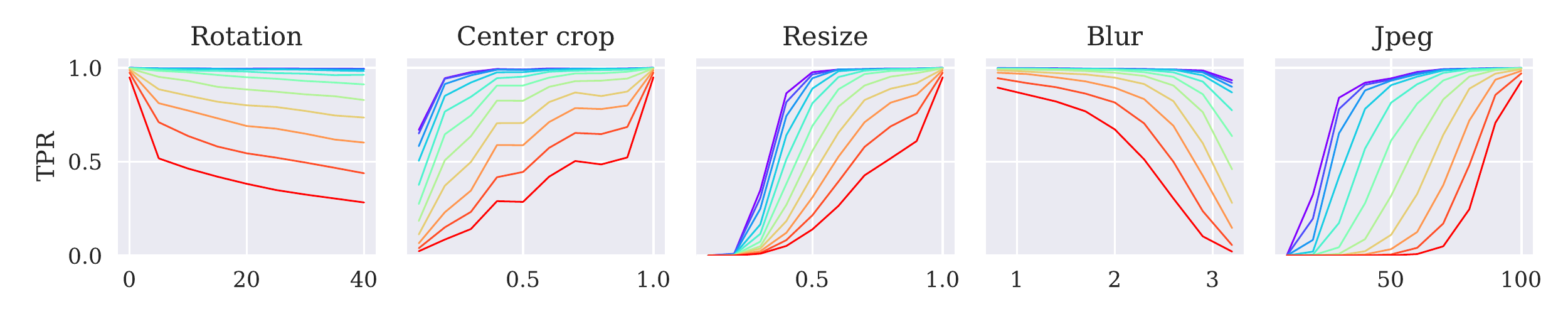}
    \end{subfigure}
    \vspace{-10pt}
    \caption{Robustness of the detection in the zero-bit setup against image transformations. \emph{Top:} PSNR set at 40~dB and FPR decreasing from \textcolor{BrickRed}{$10^{-2}$} (\textcolor{BrickRed}{red}) to \textcolor{Violet}{$10^{-12}$} (\textcolor{Violet}{blue}). \emph{Bottom:} FPR set at $10^{-6}$ and PSNR ranging from \textcolor{BrickRed}{52\,dB} to \textcolor{Violet}{32\,dB}.
    }
    \label{fig:0bit_trade_offs}
\end{figure*}
\fi

\head{Embedding}
We first set a desired FPR (which defines the hypercone angle $\theta$) and a target PSNR.
Watermarking~\eqref{eq:Watermarking} then uses the Adam optimizer~\cite{kingma2015adam} with learning rate $0.01$ over $100$ gradient descent iterations.
The weight in~\eqref{eq:global_loss} is set to $\lambda=1$ (zero-bit) or $\lambda=5\cdot 10^4$ (multi-bit). 
The margin of (\ref{eq:loss_multibit}) is set to $\mu=5$.

At each iteration, the preprocessing step performs the SSIM attenuation and clips the PSNR to the target. 
SSIM heatmaps are computed with $C_1=0.01^2$, $C_2=0.03^2$ and over $17\times 17$ tiles of the image's channels, then summed-up and clamped to be non negative, which generates a single heatmap per image. 
Then, a transformation $t$ is chosen randomly in $\mathcal T$ (identity, rotation, blur, crop or resize). 
The rotation angle $\alpha$ is sampled from a Von Mises distribution with $\mu=0$, $\kappa=1$ and divided by $2$. 
This generates angles in $\pi/2\times[-1,1]$ with a higher probability for small rotations, that are more frequent in practice. 
The crop and resize scales are chosen uniformly in $[0.2, 1.0]$. 
The crop aspect ratio is also randomly chosen between $3/4$ and $4/3$.
The blurring kernel size $b$ is randomly drawn from the odd numbers between $1$ and $15$ and $\sigma$ is set to $0.15 b+0.35$.
Finally, the image is flipped horizontally with probability $0.5$.

\head{Transformations} The following table presents the transformations used at pre-training, marking or evaluation stages. 
Parameter $p$ represents the cropping ratio in terms of area, $Q$ is the quality factor of the compression and $B$, $C$, $H$ are defined in~\cite{marcel2010torchvision}. "Meme format" and "Phone screenshot" come from the Augly library~\cite{bitton2021augly}. %

\noindent
\resizebox{1.0\linewidth}{!}{
\begin{tabular}{ l c|cc|cc }
    \toprule
                     &                  & \multicolumn{2}{c|}{Type} & \multicolumn{2}{c}{Used for}\\
    Transformations  & Parameter        & Geom & Value   & Train & Mark\\ 
    \midrule
    Rotation         & angle $\alpha$    & \cmark & \xmark      & \cmark & \cmark \\
    Crop             & ratio $p$         & \cmark & \xmark      & \cmark & \cmark \\
    Resize           & scale $p$         & \cmark & \xmark      & \cmark & \cmark \\
    Gaussian Blur    & width $\sigma$    & \xmark & \cmark      & \cmark & \cmark \\
    Brightness       & $B$               & \xmark & \cmark      & \cmark & \xmark \\
    Contrast         & $C$               & \xmark & \cmark      & \cmark & \xmark \\
    Hue              & $H$               & \xmark & \cmark      & \cmark & \xmark \\
    JPEG             & quality $Q$       & \xmark & \cmark      & \xmark & \xmark \\
    Meme format      & -                 & \cmark & \cmark      & \xmark & \xmark \\
    Phone screenshot & -                 & \cmark & \cmark      & \xmark & \xmark \\
    \bottomrule
\end{tabular}
}

\ifarxiv
\vspace{1em}
\fi

\vspace{-1em}
\subsection{Zero-bit watermarking}
\vspace{-0.5em}

\head{Trade-offs}
The hypercone angle $\theta$ in~\eqref{eq:FPR} is given by the target FPR. %
A higher FPR implies a wider angle, making the method robust against more severe attacks, at the cost of detecting more false positives. 
The FPR is set to $10^{-6}$ in further experiments.
Large-scale applications usually operate at low FPR to avoid human verification.
As a sanity check we run detection on 100k natural images from YFCC, none of which are found to be marked. 
Similarly, there is only one false positive out of the 1,281,167 images of ImageNet.
Another trade-off lies in the imperceptibility, since allowing greater distortions (lower PSNR) improves the robustness.
It is illustrated in Fig.~\ref{fig:0bit_trade_offs}.

\ifnotarxiv
\begin{figure}[h]
    \centering
    \begin{subfigure}[b]{\linewidth}
        \includegraphics[width=\linewidth, trim={0 1em 0 0em},clip]{figs/0bit_fprs.pdf}
    \end{subfigure}
    \vspace{1em}
    \begin{subfigure}[b]{\linewidth}
        \includegraphics[width=\linewidth, trim={0 0em 0 1.6em},clip]{figs/0bit_psnrs.pdf}
    \end{subfigure}
    \vspace*{-2em}
    \caption{Robustness against image transformations. \emph{Top:} PSNR set at 40~dB and FPR decreasing from \textcolor{BrickRed}{$10^{-2}$} (\textcolor{BrickRed}{red}) to \textcolor{Violet}{$10^{-12}$} (\textcolor{Violet}{blue}). \emph{Bottom:} FPR set at $10^{-6}$ and PSNR ranging from \textcolor{BrickRed}{52\,dB} to \textcolor{Violet}{32\,dB}.
    }
    \label{fig:0bit_trade_offs}
\end{figure}
\fi

\begin{figure}[b!]
    \centering
    \includegraphics[width=1.0\linewidth,clip,trim={0 0em 15pt 0em}]{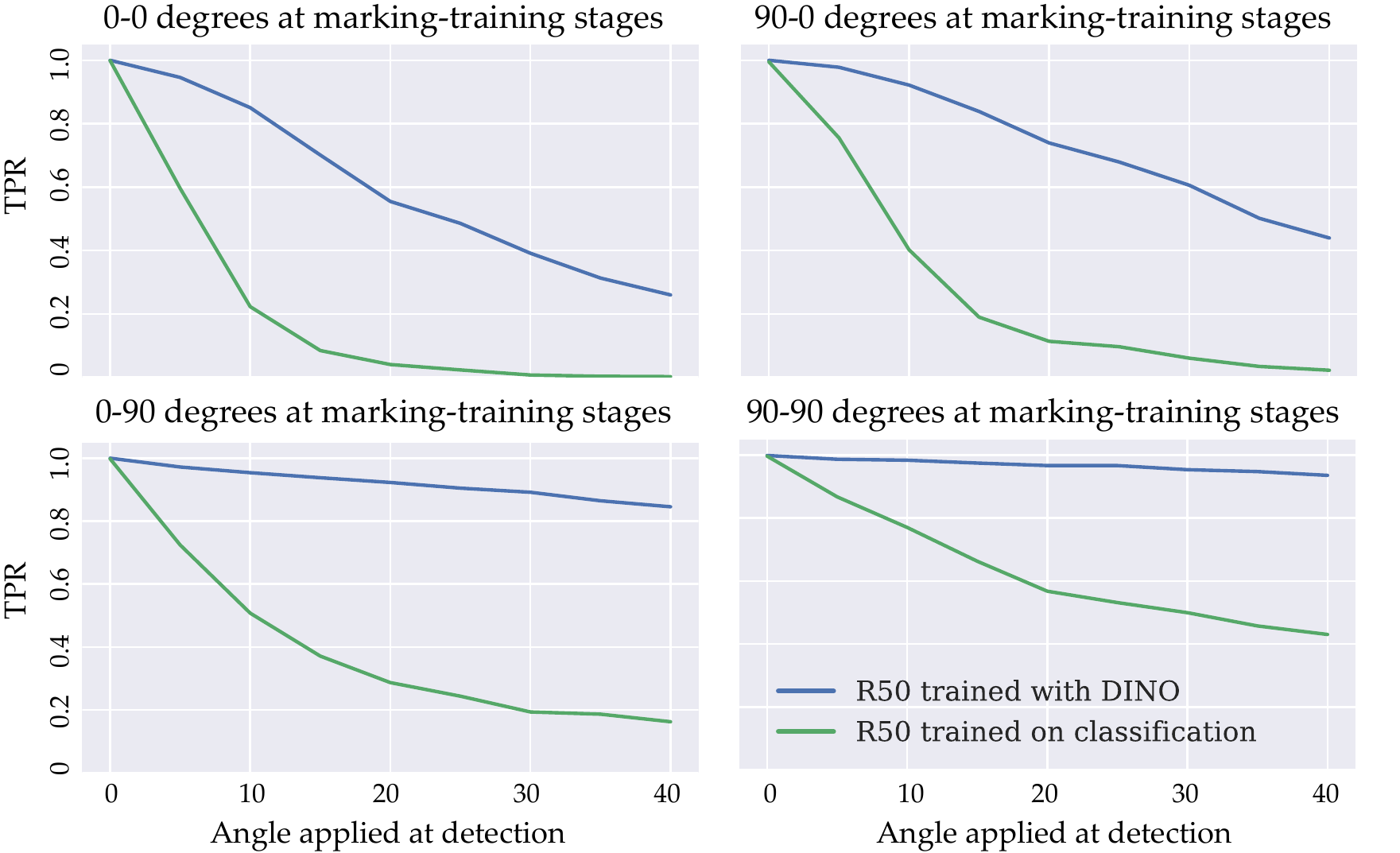}
    \caption{Robustness against rotation. Each row (column) represents different $\pm$amplitude of the rotation at training (resp. marking~\eqref{fig:0bit_rotations}). 
    }
    \label{fig:0bit_rotations}
\end{figure}

\head{Ablation studies}\label{par:0bit_quantitative_results}
We showcase the influence of self supervision at pre-training and of augmentation at marking time.
The performance measure is the True Positive Rate (TPR), at a target PSNR=40\,dB, and FPR=$10^{-6}$.
Fig.~\ref{fig:0bit_rotations} evaluates the robustness for the specific case of the rotation. 
Rotation augmentation is needed both at pre-training and marking stages to achieve high robustness against it.
Comparison on a wider range of transformations is given in Tab. \ref{tab:0bit}.

\begin{figure}[b]
    \vspace{-5pt}
    \centering
    \includegraphics[width=\linewidth]{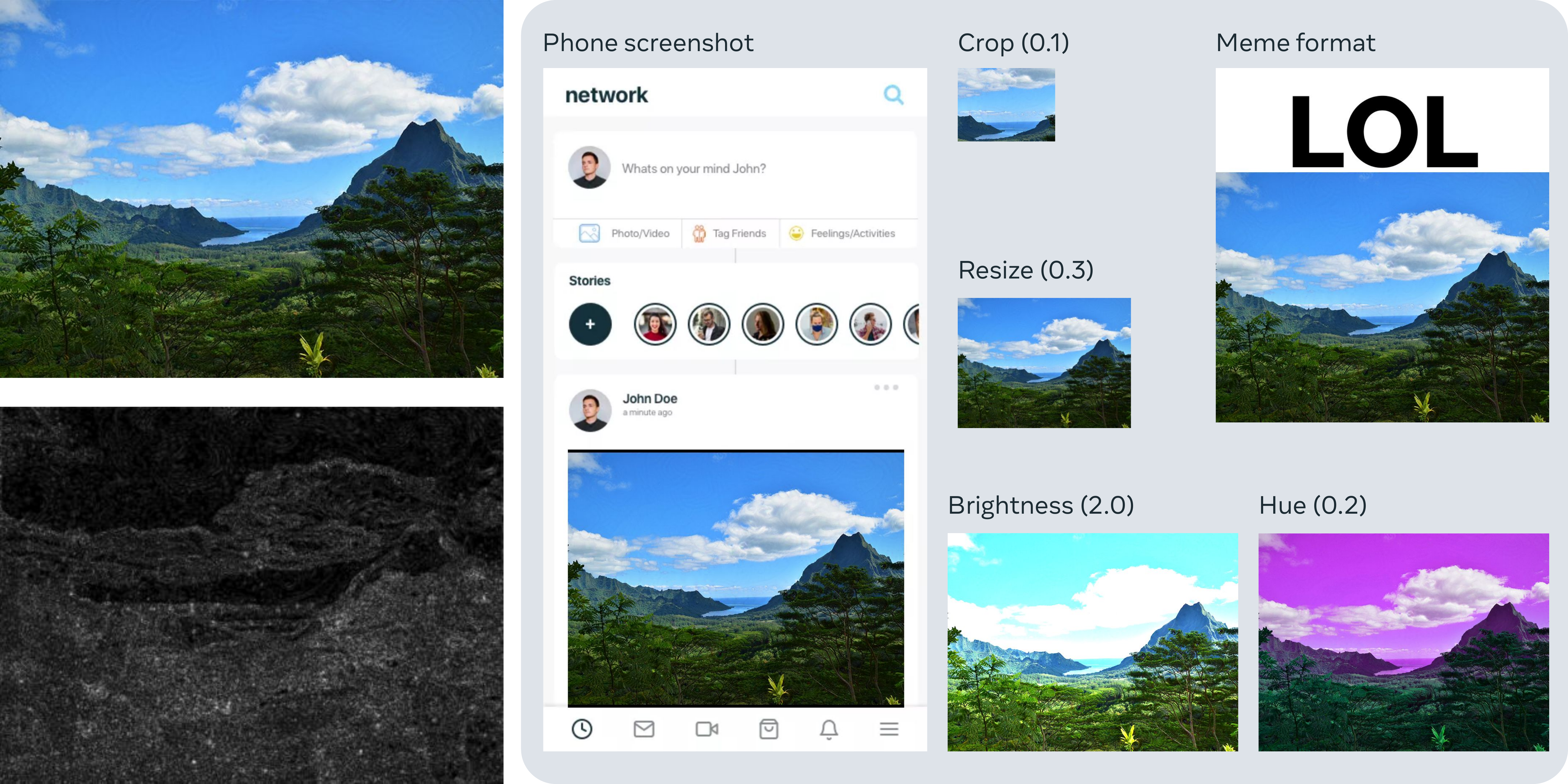}
    \vspace{-17pt}
    \caption{Example of an image ($800\times 600$) watermarked at PSNR 40~dB and FPR=$10^{-6}$, and some detected alterations. The black and white picture shows the scaled amplitude of the watermark signal.
    \label{fig:watermarked_imgs}}
    \vspace{-3pt}
\end{figure}

\head{Qualitative results}\label{par:0bit_qualitative_results}
Fig.~\ref{fig:watermarked_imgs} presents an image watermarked at PSNR=40~dB and some detected alterations, as well as its pixel-wise difference w.r.t. the original image. The watermark is almost invisible even to the trained eye because it is added in the textured regions due to the perceptual SSIM normalization applied during watermarking.

\head{Comparison with the state of the art} \autoref{tab:0bit} compares our method with~\cite{vukotic2020classification} on CLIC. 
In their setup, the FPR=$10^{-3}$ and PSNR must be $\geq 42$dB.
Overall, our method gives better results on CLIC than on YFCC because images have higher resolutions (hence more pixels can be used to convey the mark). We observe a strong improvement over \cite{vukotic2020classification}, especially for large rotations, crops and Gaussian blur where our method yields almost perfect detection over the 118 images. 

\begin{table}[t]
    \centering
    \caption{TPR over various attacks.
    1\textsuperscript{st} setup: performance with SSL vs supervised ResNet-50 networks on YFCC. 
    2\textsuperscript{nd} setup: evaluation over CLIC.
    ($\star$) best results in~\cite{vukotic2020classification} (VGG-19 trained on hard attacks and RMAC aggregation), ($\star \star$) our implementation of \cite{vukotic2020classification} (with default pre-trained VGG-19). \newline
    $\dagger$ denotes augmentations used at pre-training. 
    }
    \label{tab:0bit}
    \vspace{-0.8em}
    \resizebox{1.0\linewidth}{!}{
    \begin{tabular}{ l | >{\columncolor{Apricot!30}}lc|>{\columncolor{Apricot!30}}lcl}
        \toprule
        \multicolumn{1}{c}{}&  \multicolumn{2}{c}{Setup 1: YFCC}    &  \multicolumn{3}{c}{Setup 2: CLIC} \\
        \cmidrule(lr){2-3} \cmidrule(lr){4-6}
        Transformations     & SSL              & Sup.               & Ours                        & \cite{vukotic2020classification} ($\star$) & \cite{vukotic2020classification} ($\star\star$) \\ \hline
        Identity            & 1.00$^\dagger$   & 1.00$^\dagger$     &   1.00$^\dagger$            & 1.0$^\dagger$            & 1.00$^\dagger$  \\
        Rotation (25)       & 0.97$^\dagger$   & 0.54$^\dagger$     &   \textbf{1.00}$^\dagger$   & $\approx 0.3$$^\dagger$ & 0.27$^\dagger$  \\
        Crop (0.5)          & 0.95$^\dagger$   & 0.79$^\dagger$     &   1.00$^\dagger$            & $\approx 0.1$$^\dagger$ & 1.00$^\dagger$  \\
        Crop (0.1)          & 0.39$^\dagger$   & 0.06$^\dagger$     &   \textbf{0.98}$^\dagger$   & $\approx 0.0$$^\dagger$ & 0.02$^\dagger$  \\
        Resize (0.7)        & 0.99$^\dagger$   & 0.85$^\dagger$     &   1.00$^\dagger$            & -                        & 1.00$^\dagger$  \\
        Blur (2.0)          & 0.99$^\dagger$   & 0.04               &   \textbf{1.00}$^\dagger$   & -                        & 0.25  \\
        JPEG (50)           & 0.81             & 0.20               &   0.97                      & $\approx 1.0$            & 0.96  \\
        Brightness (2.0)    & 0.94$^\dagger$   & 0.71               &   0.96$^\dagger$            & -                        & 0.99  \\
        Contrast (2.0)      & 0.96$^\dagger$   & 0.65               &   1.00$^\dagger$            & -                        & 1.00  \\
        Hue (0.25)          & 1.00$^\dagger$   & 0.46               &   1.00$^\dagger$            & -                        & 1.00  \\
        Meme                & 0.99             & 0.94               &   1.00                      & -                        & 0.98  \\
        Screenshot          & 0.76             & 0.18               &   \textbf{0.97}             & -                        & 0.86  \\
        \bottomrule
    \end{tabular}
    }
    \vspace{-1.0em}
\end{table}

\vspace{-1em}
\subsection{Multi-bit data hiding}\label{sec:multibit_exp}
\vspace{-0.5em}

\head{Quantitative results}\label{par:multibit_quantitative_results}
We evaluate the method on YFCC, with  a target PSNR of 40~dB and a payload $k$ of $30$ random bits as in~\cite{zhu2018hidden,luo2020distortion}.
Tab.~\ref{tab:multibit_30} presents the Bit and Word Error Rate (BER and WER) over various attacks. The decoding achieves low rates over a wide range of geometric (rotation, crops, resize, etc.) and valuemetric (brightness, hue, contrast, etc.) attacks.
Rotation and Gaussian blur are particularly harmless since they are seen both at pre-training and at marking time.
Some images are harder to mark, which can be observed statistically on the empiricial WER reported in Tab.~\ref{tab:multibit_30}. 
If all images were as difficult to mark, then BER would be equal for all images, and $\mathrm{WER} =1- (1-\mathrm{BER})^{k}$. 
Yet, the reported WER are significantly lower: \eg for Brigthness, $\mathrm{WER} = 0.607 < 1-(1-0.087)^{30} = 0.935$. 
Empirically, we see that images with little texture are harder to watermark than others, due to the SSIM normalization.
In practice, ECC can be used to achieve lower WERs.

\soutx{\begin{figure}[h!]
    \centering
    \includegraphics[width=\textwidth]{figs/multibit_30.pdf}
    \caption{Average bit accuracy (blue) and word accuracy (green) of the decodings, over 1000 images of the YFCC dataset, against the strength of the transformations applied before detection.
    The target PSNR is set to $40$, i.e. all the watermarked images have PNSR $\geq 40$, the encoded messages have length $30$.}
    \label{fig:multibit_30}
\end{figure}}

\begin{table}[t]
    \centering
    \caption{\makebox{BER and WER (\%) for $30$-bits encoding at PSNR~40dB.}}
    \label{tab:multibit_30}
    \vspace{-10pt}
    \resizebox{1.0\linewidth}{!}{
    \begingroup
        \setlength{\tabcolsep}{3pt}
            \begin{tabular}{ c| *{12}{c}}
            \multicolumn{1}{c}{\rot{Transform.}} & \rot{Identity} & \rot{Rot. (25)} & \rot{Crop (0.5)} & \rot{Crop (0.1)} & \rot{Res. (0.7)} & \rot{Blur (2.0)} & \rot{JPEG (50)} & \rot{Bright. (2.0)} & \rot{Contr. (2.0)} & \rot{Hue (0.25)} & \rot{Meme} & \rot{Screenshot} \\ \midrule
            BER  & 0.1   & 3.3    & 4.8    & 28.9   & 2.1    & 0.5  & 20.8   & 8.7   & 8.4   & 2.5   & 6.4    & 23.9  \\
            WER  & 0.7   & 43.4   & 58.3   & 100    & 29.1   & 4.6  & 98.9   & 60.7  & 62.6  & 31.6  & 75.9   & 100   \\
            \bottomrule
    \end{tabular}
    \endgroup
    }
    \ifnotarxiv
    \vspace{-1em}
    \fi
\end{table}

\head{Qualitative results}
We notice that the watermark is perceptually less visible for multi-bit than for zero-bit watermarking at a fixed PSNR. Our explanation is that the energy put into the image feature is more spread-out across carriers in the multi-bit setup than on the zero-bit one where the feature is pushed at much as possible towards a single carrier.
Images are not displayed due to lack of space.

\soutx{\begin{figure}[h!]
    \centering
    \includegraphics[width=0.3\textwidth]{figs/imgs/multibit_psnr40/0_out.png}
    \includegraphics[width=0.3\textwidth]{figs/imgs/multibit_psnr40/1_out.png}
    \includegraphics[width=0.3\textwidth]{figs/imgs/multibit_psnr40/2_out.png}
    \includegraphics[width=0.3\textwidth]{figs/imgs/multibit_psnr40/0_diff_30.png}
    \includegraphics[width=0.3\textwidth]{figs/imgs/multibit_psnr40/1_diff_30.png}
    \includegraphics[width=0.3\textwidth]{figs/imgs/multibit_psnr40/2_diff_30.png}
    \caption{Example of watermarked images at PSNR=40 (Top row). The second row gives the pixel-wise $l_1$ distance between the images and their watermarked counterpart. The messages to encode are 30 bits long.}
    \label{fig:watermarked_imgs_multibit}
\end{figure}}

\head{Comparison with the state of the art}
\autoref{tab:multibit_coco} compares against two deep data hiding methods~\cite{zhu2018hidden,luo2020distortion} using their setting: a payload of $30$ bits, a target PSNR of 33dB, over $1000$ images from COCO resized to $128 \times 128$. 
Overall, our method gives comparable results to the state of the art, except for the center crop transform where it fails to achieve high bit accuracies. 
Note also that the resize operation is not used as noise layer in neither of~\cite{zhu2018hidden,luo2020distortion}, which means that our method should have the advantage.
In contrast, while these methods are trained to be robust to JPEG compression with a differentiable approximation, our method achieves similar performance without specific training.

Furthermore, our method easily scales to higher resolution images, where it achieves lower BER for a fixed payload. 
We assume that \cite{zhu2018hidden,luo2020distortion} also scales but at the cost of a specific training for a given resolution. 
This training is more computationally expensive since the message is repeated at each pixel~\cite{zhu2018hidden}. 
It also needs a smaller batch-size to operate larger images, and new hyperparameters values.

\ifarxiv
\begin{table}[t]
    \centering
    \caption{Comparison of BER. The first row uses original resolutions of COCO, while the others use a resized version (to $128 \times 128$). Results for \cite{zhu2018hidden, luo2020distortion} come from \cite{luo2020distortion}. $\dagger$ denotes transformations used in the watermarking process.
    }
    \label{tab:multibit_coco}
    \vspace{-1.5em}
    \resizebox{1.0\linewidth}{!}{
    \begin{tabular}{ c|llllll}
        \multicolumn{1}{c}{Transformation} & \rotatebox{70}{Identity} & \rotatebox{70}{JPEG (50)} & \rotatebox{70}{Blur (1.0)} & \rotatebox{70}{Crop (0.1)}  & \rotatebox{70}{Resize (0.7)} & \rotatebox{70}{Hue (0.2)} \\ \midrule
        \rowcolor{Orchid!20} Ours                      & 0.00$^\dagger$ & 0.04           & 0.00$^\dagger$          & 0.18$^\dagger$            & 0.00$^\dagger$ & 0.03          \\ \midrule
        \rowcolor{Apricot!30} Ours, $128 \times 128$                          & 0.00$^\dagger$ & 0.16  & 0.01$^\dagger$ & 0.45$^\dagger$            & 0.18$^\dagger$ & 0.06 \\
        HiDDeN~\cite{zhu2018hidden}                     & 0.00$^\dagger$ & 0.23$^\dagger$ & 0.01$^\dagger$ & 0.00$^\dagger$   & 0.15           & 0.29          \\
        Dist. Agnostic~\cite{luo2020distortion}         & 0.00$^\dagger$ & 0.18$^\dagger$ & 0.07$^\dagger$          & 0.02$^\dagger$            & 0.12  & 0.06 \\
        \bottomrule
    \end{tabular}}
    \vspace{-0.7em}
\end{table}
\fi

\ifnotarxiv
\begin{table}[h]
    \centering
    \caption{Comparison of BER. The first row uses original resolutions of COCO, while the others use a resized version (to $128 \times 128$). Results for \cite{zhu2018hidden, luo2020distortion} come from \cite{luo2020distortion}. $\dagger$ denotes transformations used in the watermarking process.
    }
    \label{tab:multibit_coco}
    \vspace{-2em}
    \resizebox{1.0\linewidth}{!}{
    \begin{tabular}{ c|llllll}
        \multicolumn{1}{c}{Transformation} & \rotatebox{70}{Identity} & \rotatebox{70}{JPEG (50)} & \rotatebox{70}{Blur (1.0)} & \rotatebox{70}{Crop (0.1)}  & \rotatebox{70}{Resize (0.7)} & \rotatebox{70}{Hue (0.2)} \\ \midrule
        \rowcolor{Orchid!20} Ours                      & 0.00$^\dagger$ & 0.04           & 0.00$^\dagger$          & 0.18$^\dagger$            & 0.00$^\dagger$ & 0.03          \\ \midrule
        \rowcolor{Apricot!30} Ours, $128 \times 128$                          & 0.00$^\dagger$ & 0.16  & 0.01$^\dagger$ & 0.45$^\dagger$            & 0.18$^\dagger$ & 0.06 \\
        HiDDeN~\cite{zhu2018hidden}                     & 0.00$^\dagger$ & 0.23$^\dagger$ & 0.01$^\dagger$ & 0.00$^\dagger$   & 0.15           & 0.29          \\
        Dist. Agnostic~\cite{luo2020distortion}         & 0.00$^\dagger$ & 0.18$^\dagger$ & 0.07$^\dagger$          & 0.02$^\dagger$            & 0.12  & 0.06 \\
        \bottomrule
    \end{tabular}}
    \vspace{-0.7em}
\end{table}
\fi

\soutx{\begin{figure}[h!]
    \centering
    \includegraphics[width=\textwidth]{figs/multibit_yfcc_coco.pdf}
    \caption{Bit accuracy against the strength of the transformations applied before detection, over 1000 images of the COCO and YFCC datasets, resized to $128\times 128$ resolution or not. The target PSNR is set to $33$, i.e. all the watermarked images have PNSR $\geq 33$, and the message length is set $30$.}
    \label{fig:multibit_yfcc_coco}
\end{figure}}

\section{Conclusion \& Discussion}

This paper proposes a way to robustly and invisibly embed information into digital images, by watermarking onto latent spaces of off-the-shelf self-supervised networks. 
By incorporating data augmentation and constraints into the marking process, our zero-bit watermarking method greatly improves performance over the baseline \cite{vukotic2020classification}.
It is robust against a wide range of transformations while keeping high fidelity with regards to the original images, and ensuring a very low false positive rate for the detection.
When we extend the method to multi-bit watermarking, we obtain promising results, comparable to the state-of-the-art in deep data hiding, and even better with regards to some transformations of the image (\eg JPEG compression or blur).

Most interestingly, networks trained with self-supervision naturally generate excellent watermarking spaces, without being explicitly trained to do so. 
However, compared to encoder-decoder deep watermarking techniques, watermarking images with our method is expansive since it is not a single pass forward. 
In future works, we hope to show that further adapting the network for the specific task of watermarking would improve performance and efficiency.

\ifarxiv
\bibliographystyle{IEEEbib}

\renewcommand{\arraystretch}{0.96}
\renewcommand{\baselinestretch}{0.96}

\fi

        \bibliography{references}

\clearpage
\onecolumn
\section*{Appendix}\label{section:Appendix}

\subsection{Demonstration of the False Positive Rate formula (Eq. \ref{eq:FPR})}

One way to sample $U$ is the following: First sample a white Gaussian vector $G \sim \mathcal{N}(0,I)$ in $\mathbb R ^d$ and then normalize: $U=G/\norm{G}$.
Without loss of generality, we can assume that $a = (1,0,...,0)$ (up to a change of axes). Therefore, $$U^T a = \frac{G_1}{\sqrt{\sum_{i=1}^d G_i^2}}$$
Let $\tau \in [0,1]$. We have 
\begin{align*}
    \mathbb{P}\left((U^T a)^{2} \geq \tau^{2}\right) &=\mathbb{P}\left(\frac{G_{1}^{2}}{\sum_{i=1}^{d} G_{i}^{2}} > \tau^{2}\right)
    =\mathbb{P}\left(\frac{G_{1}^{2}}{\sum_{i=2}^{d} G_{i}^{2}} > \frac{\tau^{2}}{1-\tau^{2}}\right)\\
    &=\mathbb{P}\left((d-1)\frac{G_{1}^{2}}{\sum_{i=2}^{d} G_{i}^{2}} > (d-1)\frac{\tau^{2}}{1-\tau^{2}}\right)
\end{align*}
Note that $Y := (d-1)\frac{G_{1}^{2}}{\sum_{i=2}^{d} G_{i}^{2}}$ is the ratio of two independent chi-squares random variables of degree 1 and $d-1$. By definition, $Y$ follows a Fisher distribution $F(1,d-1)$. Its cumulative density function is: $F(x; 1, d-1) = I_{x/(x+d-1)}(1/2, (d-1)/2)$. 

It follows, using $x=(d-1)\frac{\tau^2}{1-\tau^2} = -(d-1) + (d-1)\frac{1}{1-\tau^2}$ and setting $\tau = \cos\theta$, that 
$$\mathbb P (\abs{U^T a} > \cos \theta) = \mathbb P (Y> \cos^2 \theta) = 1 - I_{\cos^2 \theta}(1/2, (d-1)/2) = I_{\sin^2 \theta}( (d-1)/2,1/2)$$
where $I_x(a,b)$ is the regularized incomplete beta function.

\subsection{Low resolution watermarked images}

\begin{figure}[h!]
    \includegraphics[width=\textwidth]{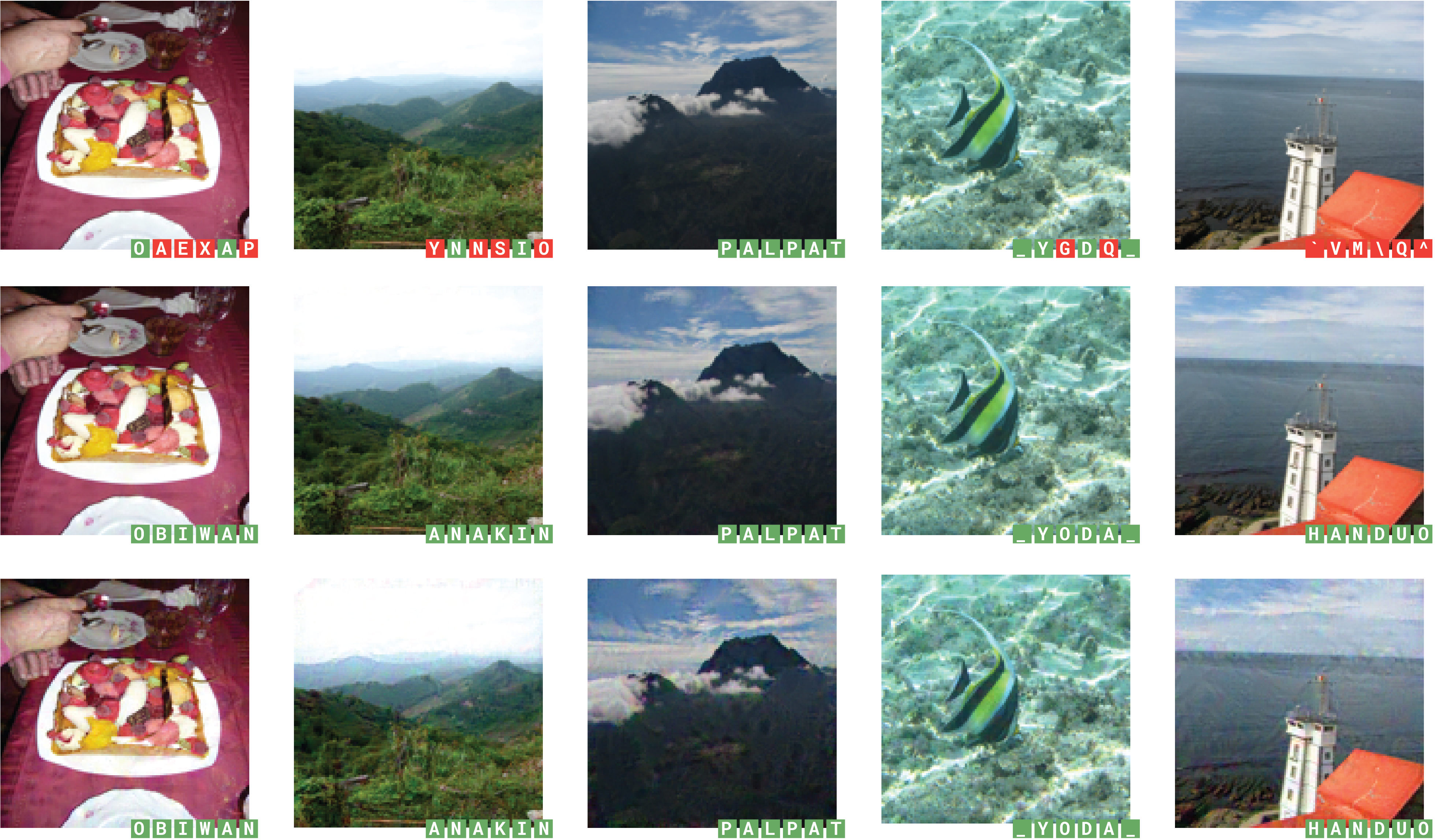}
    \caption{Watermarked images from the INRIA Holidays dataset resized to $128 \times 128$. The watermark is added with our multi-bit watermarking method with a payload of $30$ bits and with different values for the target PSNR: $52$dB (top row), $40$dB (middle row), $32$dB (bottom row). 
    We use a 5-bits character encoding to encode and decode one message per image, and show the decoded messages for each image (without any transformation applied to the image). The higher the PSNR, the higher the decoding errors, and the less robust the decoding is to transformations.}
    \label{fig:holidays_resized_multibit}
    \vspace{-2cm}
\end{figure}

\newpage
\subsection{High resolution watermarked images}

\begin{figure}[h!]
     \centering
     \begin{subfigure}[b]{0.95\textwidth}
         \includegraphics[width=\textwidth]{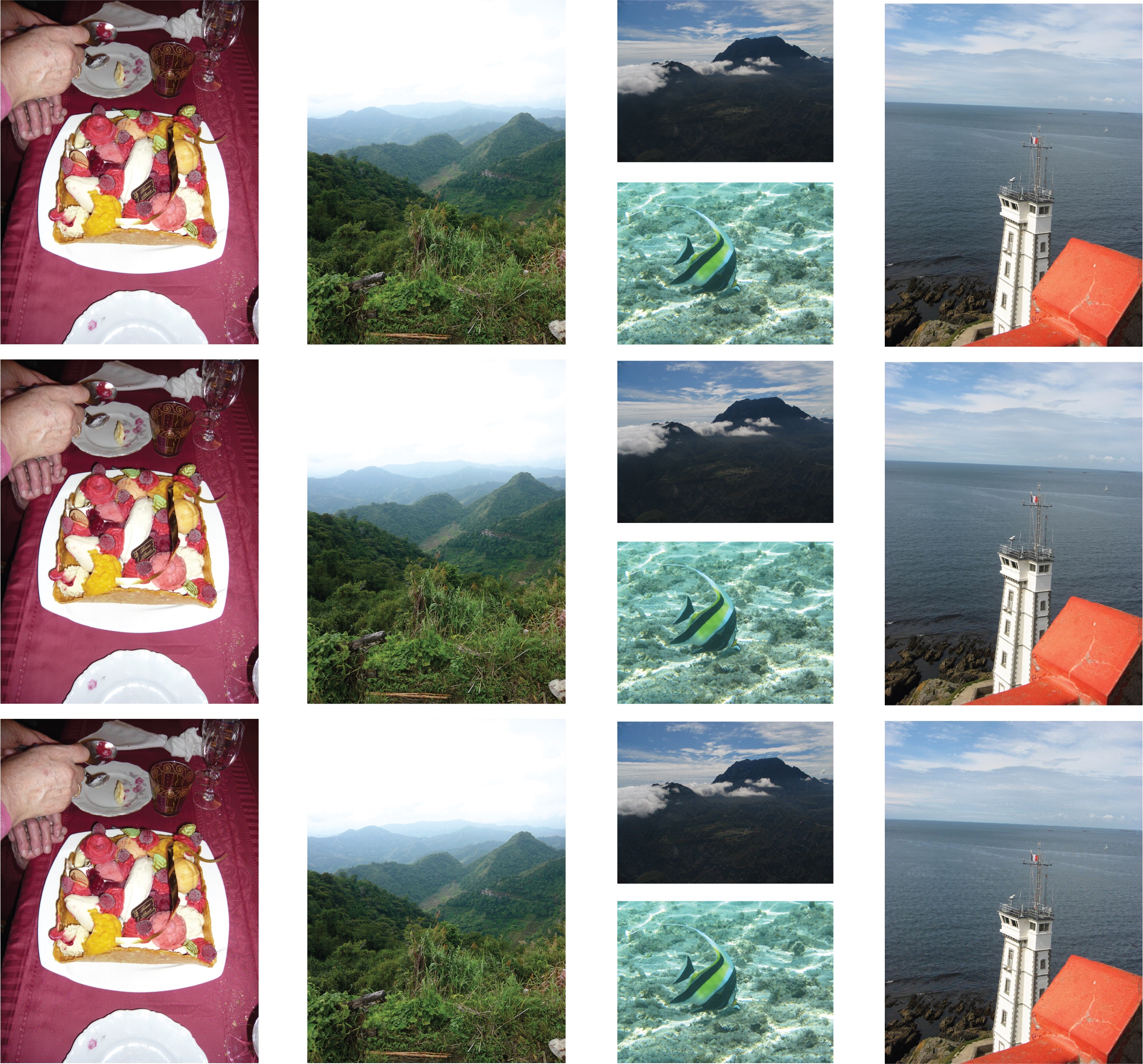}
         \caption{The watermark is added with our zero-bit watermarking method with an FPR of $10^{-6}$ and with different values for the target PSNR: $52$dB (top row), $40$dB (middle row), $32$dB (bottom row)}
         \label{fig:holidays_0bit}
     \end{subfigure}
     \begin{subfigure}[b]{0.95\textwidth} \vspace{0.2cm}
         \includegraphics[width=\textwidth]{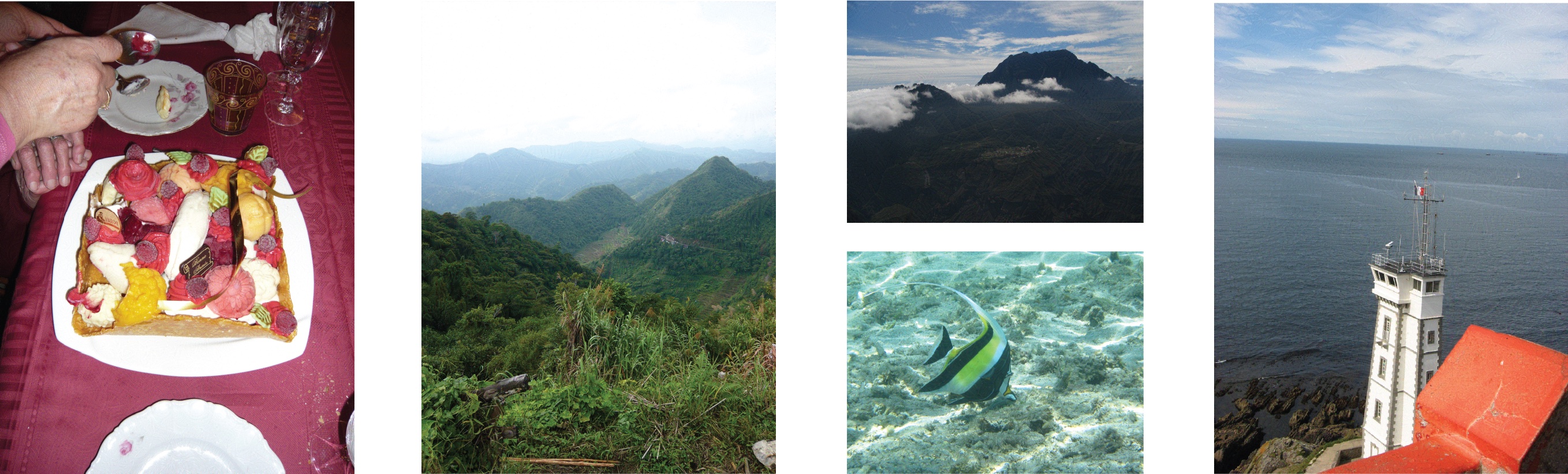}
         \caption{The watermark is added with our multi-bit watermarking method with a payload of $30$ bits with the target PSNR at $32$dB.}
         \label{fig:holidays_multibit}
     \end{subfigure}
        \caption{Watermarked images from the INRIA Holidays dataset (resolution around $2048 \times 1536$).}
        \label{fig:holidays}
        \vspace{-4.0cm}
\end{figure}

\end{document}